\documentclass[conference]{IEEEtran}
\IEEEoverridecommandlockouts
\usepackage{cite}
\usepackage{amsmath,amssymb,amsfonts}
\usepackage{algorithmic}
\usepackage{graphicx}
\usepackage{textcomp}
\usepackage{xcolor}

\usepackage[hyphens]{url}
\def\BibTeX{{\rm B\kern-.05em{\sc i\kern-.025em b}\kern-.08em
    T\kern-.1667em\lower.7ex\hbox{E}\kern-.125emX}}

\begin{document}

\title{Evaluating Multimodal LLMs 
across Text and Audio Modalities for Accessible Disaster Assistance}

\author{
\IEEEauthorblockN{Anuridhi Gupta}
\IEEEauthorblockA{\textit{Humanitarian Informatics Lab}\\ \textit{George Mason University}\\
Fairfax, VA, USA\\
agupta29@gmu.edu}
\and
\IEEEauthorblockN{Samara Mansoor}
\IEEEauthorblockA{\textit{Humanitarian Informatics Lab}\\ 
\textit{George Mason University}\\
Fairfax, VA, USA\\
smansoo5@gmu.edu}
\and
\IEEEauthorblockN{Hemant Purohit}
\IEEEauthorblockA{\textit{Humanitarian Informatics Lab}\\ 
\textit{George Mason University}\\
Fairfax, VA, USA\\
hpurohit@gmu.edu}
}

\maketitle

\begin{abstract}
Effective disaster risk communication is a foundational humanitarian challenge, yet current emergency infrastructure fails to meet the needs of individuals with access and functional needs, including hard-of-hearing individuals, pregnant women, mothers with toddlers, and elderly individuals with dementia. Recent advancements in Artificial Intelligence (AI), especially Multi-Modal Large Language Models (MM-LLMs), demonstrate powerful capabilities to serve diverse users across text, audio, image, and video modalities within a single unified system, such as a chatbot. However, their suitability for deployment rests on a property that receives limited scrutiny, i.e., whether these systems produce consistent, actionable outputs regardless of the modality through which a user communicates. In this paper, we conduct a comprehensive analysis to understand the status of open-weight MM-LLMs using real emergency alert scenarios across four different vulnerable personas.  These state-of-the-art (SOTA) models are evaluated on consistency of responses across text and audio modalities when the same task scenario is given. Findings indicate that no model achieves reliable consistency across modalities, and that performance gaps are heightened for personas with access needs, introducing modality-dependent inequity that undermines the humanitarian value of these systems. These results inform concrete design recommendations for building equitable, trustworthy, and inclusive AI tools for disaster risk communication.

\end{abstract}



\section{Introduction}
Natural disasters 
represent 
diverse emergency situations faced by communities worldwide. For instance, from earthquakes and storms to floods and droughts, the U.S. sustained 403 weather and climate disasters from 1980 to 2024 
with costs over 1 billion dollars~\cite{smith20242023}. The frequency and intensity of such events continues to rise under shifting climate conditions~\cite{smith20242023}.
Thus, to enable the public better prepare, respond to, and recover from disasters, 
there has been a noticeable emphasis on effective disaster risk communication~\cite{stewart2024advancing}. This includes early warning devices and alerting systems that can monitor, forecast, and 
communicate risks to enable individuals, 
communities, governments, businesses and others to take timely action to reduce disaster risks
before hazardous events occur~\cite{itu2025q31}. Such risk communication systems are increasingly being modernized nowadays, including experimenting with 
Artificial Intelligence (AI) techniques~\cite{urbanelli2024ermes,foubert2026investigating}. 

While the general public faces significant risks during these disasters, there are certain groups of people who are disproportionately affected by them and may require additional  accommodation in the design of risk communication systems. Vulnerability refers to the characteristics of a person or group and their situation that influences their ability to prepare, cope, respond, and recover from a disaster~\cite{rivera2021disaster}. The vulnerable population is less likely to have access to resources and may encounter prejudice, discrimination, and stigma due to their socio-economic status, race/ethnicity, gender, age, cognitive and/or physical ability, etc~\cite{benevolenza2019impact}. According to the U.S. Federal Emergency Management Agency (FEMA), people with access and functional needs make up to 43\% of the U.S. 
population and may increase as a result of a
disaster~\cite{fema2024afn}. People with `access and functional needs' include individuals with disabilities, individuals with limited English proficiency, individuals with limited access to transportation, individuals with limited access to financial resources, older adults, etc.~\cite{cdc2021afn}. 

Risk communication tools to assist and support the vulnerable population 
require provisions of information accessibility that can create and embed effective means of interaction for these individuals with various needs and preferences~\cite{lambert2025enhancing,cdc2021afn}. 
While 
AI systems have shown promising applications for disaster risk management, in order to 
better understand 
the situations 
around us for risk communication,  
AI needs to interpret and reason about the diverse interaction needs of individuals with 
access and functional needs~\cite{baltruvsaitis2018multimodal}. To bridge this gap between human perception and AI, multi-modal 
interfaces seek to leverage natural human capabilities to communicate via speech, gesture, touch, facial expression, and other modalities, resulting in more refined interaction~\cite{turk2014multimodal}. One such advancement is multi-modal Large Language Models (MM-LLMs)~\cite{yin2024survey}, which are able to take inputs, process, and generate outputs across 
text, image, audio, and video. 

\begin{figure*}[t]
    \centering
    \includegraphics[width=0.85\textwidth]{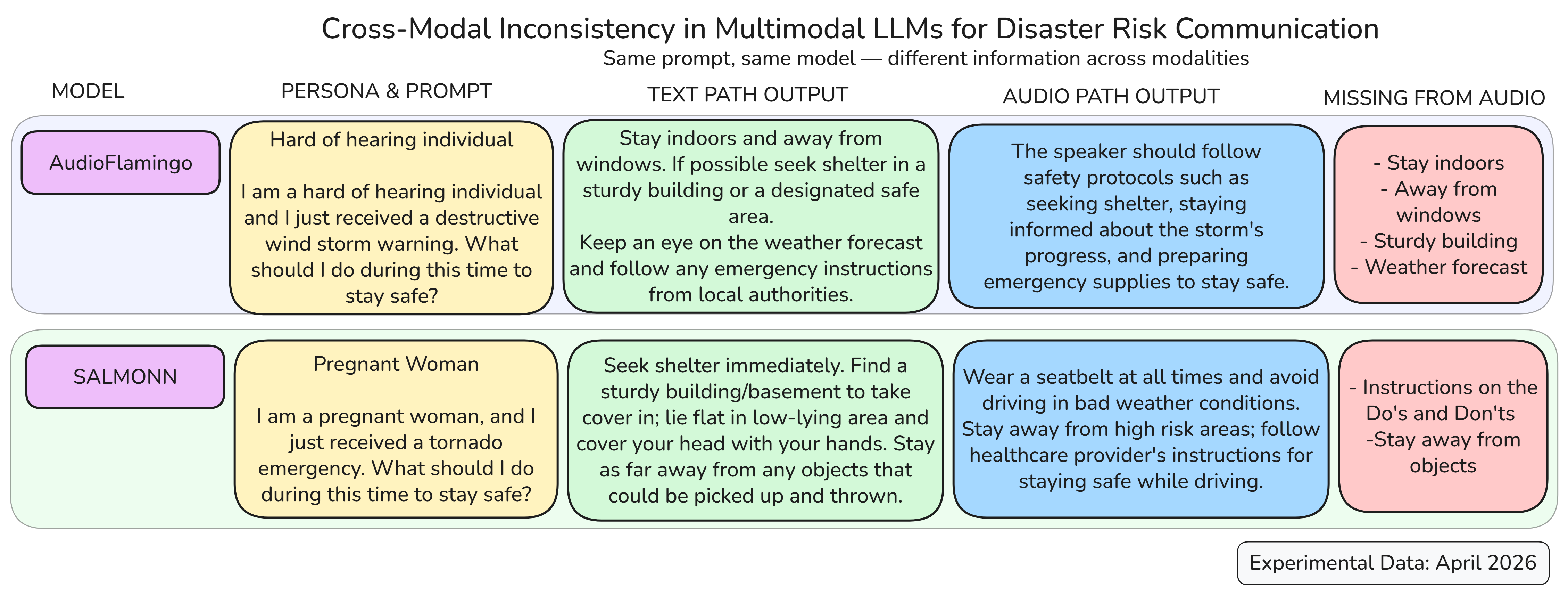}
    \caption{Illustration of cross-modal inconsistency in AudioFlamingo and SALMONN model where the same emergency query submitted via text 
    and audio modalities receives a detailed, actionable text response 
    and a generic.} 
    \label{fig:problem}
\end{figure*}

As we progress towards employing the MM-LLMs based AI technologies for public communication,  their evaluation plays a crucial role in development and deployment. Despite the claim of achieving a unified multimodal  
system design, these systems often demonstrate misalignment in understanding and response generation across input modalities~\cite{zhao2025unified}. Fig.~\ref{fig:problem} illustrates this problem. If not carefully designed and deployed, they are prone to hallucinations, are fragile to adversarial input, and fail to produce appropriate responses in high-stake domains, such as neglecting the urgency of the risk and accessibility constraints in disaster risk communication. Prior work has suggested that some of these issues correlate with the inconsistency of LLMs, which is generally defined as their tendency to generate low-confidence responses or conflicting responses when the same input prompt is resampled~\cite{wu2025estimating}. Accurately estimating MM-LLM consistency is important during critical applications such as disaster risk communication and affects 
the user's level of trust in the AI-powered systems~\cite{liang2024hemm}.

To support the diverse communication requirements of people with access and functional needs, MM-LLMs based systems must provide multiple modes of interaction effectively: voice, text, or video, while maintaining consistent interpretation across these modalities 
during communication. 
This paper proposes an evaluation framework that serves as the first step in the design 
of responsible AI systems based on consistent MM-LLMs 
for disaster risk communication. Specifically, we make the following contributions in this paper: 
\begin{enumerate}
    \item This study introduces a consistency evaluation framework 
for open-weight MM-LLMs, incorporating both  quantitative and qualitative analyses, with an emphasis on supporting multimodal interaction requirements in real-world disaster risk communication scenarios.
    \item In addition to assessing current capabilities, this work evaluates variations in MM-LLM behavior 
across a spectrum of functional and access needs, thus measuring inclusivity in model responses if deployed for risk communication.   These findings have broad implications for advancing equitable, trustworthy, and human-centered AI systems that enhance accessibility and resilience in critical communication contexts.

\end{enumerate}

The rest of the paper is structured as follows. Section~\ref{sec:related} presents the related work, followed by Section~\ref{sec:methodology}, a description of the methodology. Section~\ref{sec:results} discusses the results of empirical analysis, followed by conclusion in Section~\ref{sec:conclusion}.

\section{Related Work}
\label{sec:related}

\subsection{Risk Communication and Supporting Technologies}
Extensive literature exists for risk communication  
during disasters that often emphasizes 
that information needs to be actionable, accessible, and appropriately tailored to the needs of diverse individuals~\cite{benevolenza2019impact,stewart2024advancing,urbanelli2024ermes}. The International Telecommunication Union (ITU) has identified 
telecommunications and early warning systems as foundational infrastructure for all phases of disaster 
risk reduction and management, emphasizing that warning systems must reach all 
populations at risk with clear and usable guidance~\cite{itu2025q31}. Despite this 
imperative, traditional broadcast-based emergency alert systems such as sirens, 
television crawls, and Wireless Emergency Alerts have well-documented limitations 
in reaching individuals with sensory, cognitive, or linguistic barriers~\cite{villarreal2025barriers}.
With the emergence of conversational AI and LLMs, there are new possibilities for 
interactive, personalized risk communication~\cite{urbanelli2024ermes}. Early work explored ChatGPT's potential 
for disaster prevention information dissemination, science education, and emergency 
response support, noting its rapid availability and natural language reasoning as 
distinctive advantages over static alert systems~\cite{xue2023application}. More 
recently, LLMs have been applied to classify crisis information from social media streams, 
monitor infrastructure during active disasters, and generate structured 
warning messages grounded in official 
guidelines~\cite{linardos2025utilizing,chen2024enhancing}. 

In this work, we systematically examine the assumption that an AI system will produce reliable, consistent outputs regardless of how a user interacts with it. 
For populations with access and functional needs, who may be unable to use text-based 
interfaces under 
disaster conditions, whether the same system would provide equivalent guidance to them is a gap that still remains. 
We address 
this 
directly by evaluating  current open-weight MM-LLMs that serve as backbone of risk communication tools across text and audio modalities. 

\subsection{MM-LLMs based Systems}
MM-LLMs based systems take a user query as input and generate the response as output in the interaction modality specified by the user. The system generally comprises three core components: the \textit{Modality Encoder} that encodes inputs from diverse modalities; \textit{LLM Backbone} that does zero-shot generalization, Chain-of-Thought (CoT), and instruction following; and \textit{Modality Generator} that produces outputs in distinct modalities~\cite{zhang2024mm}. 
This architecture allows MM-LLMs to process and generate content spanning text, image, audio, and video within a single unified system, representing a significant 
improvement from earlier task-specific models that operated on single modalities~\cite{yin2024survey}.
However, a critical design challenge in MM-LLMs remains modality alignment, which ensures representations from different input modalities are mapped into a shared semantic space that the LLM backbone can reason over uniformly. Recent advances address this challenge by employing lightweight adapter modules, such as Q-Formers and linear projections, 
that bridge pre-trained modality-specific encoders to a frozen or partially-tuned 
LLM~\cite{wu2024next}. This approach has resulted in a diverse ecosystem of models, including vision-based models such as BLIP-2, 
Flamingo
, MiniGPT-4
and LLaVA
, as well as audio-vision-based models such as Video-LLaMA~\cite{zhang2023video} and SALMONN~\cite{tang2023salmonn}, which pair pre-trained speech encoders like Whisper with LLM backbones via alignment layers. 
More recent MM-LLMs are now capable of processing any combination of 
modality without cascaded pipelines~\cite{zhang2024mm}. These models employ multi-stage progressive alignment 
during training by first aligning individual modality pairs before jointly fine-tuning across modalities. Some examples include Qwen3-Omni~\cite{xu2025qwen3} and 
Audio Flamingo~\cite{goel2025audio}. 

While the above examples of models 
show promising strategies to improve 
efficiency, 
there exists a risk of cross modality interference, 
where training on one modality degrades performance or consistency in another~\cite{jiang2025specific}. This is seen in omni-modal LLMs where model attends to dominant modalities, typically text, while producing degraded or inconsistent outputs for audio inputs~\cite{jiang2025specific}. 
This structural asymmetry is the core motivation for our evaluation that despite being presented as unified systems, the underlying training dynamics of MM-LLMs 
provide opportunities 
to expect inconsistency across modalities in response generation, particularly for 
underrepresented input types such as audio data 
in safety-critical contexts like disaster risk communication. 

\subsection{Evaluation of MM-LLMs}
Current evaluation of MM-LLMs relies primarily on benchmarking MM foundational models~\cite{liang2024hemm}. Advances in both text-based and audio/vision-based MM-LLMs have led to a new set of benchmarks designed to track and guide their development efficiently~\cite{wang2025audiobench}.  These benchmarks can be broadly categorized based on text, vision, and audio-based LLMs. Text-based LLMs are evaluated on their reasoning capabilities, correct answer generation, and possible mitigation of bias~\cite{gao2025llm}. Vision LLMs focus on attack, hallucination, ethical, and cultural aspects or input modalities, i.e., visual or language perspective~\cite{tu2024many}. However, benchmarks built on modality or task-specific datasets are increasingly inadequate for capturing general capabilities~\cite{liang2024hemm}. Recent research explores human-like any-to-any modality conversion as a step toward artificial general intelligence~\cite{zhang2024mm}, that would require further comprehensive benchmarking. However, these existing works still neglect understanding whether MM-LLMs provide consistent responses across modalities for the same domain-specific task, such as the facilitation of  
risk communication for user queries. 



\section{Methodology: Consistency Evaluation Framework}
\label{sec:methodology}
\begin{figure*}[t]
    \centering
    \includegraphics[width=0.85\textwidth]{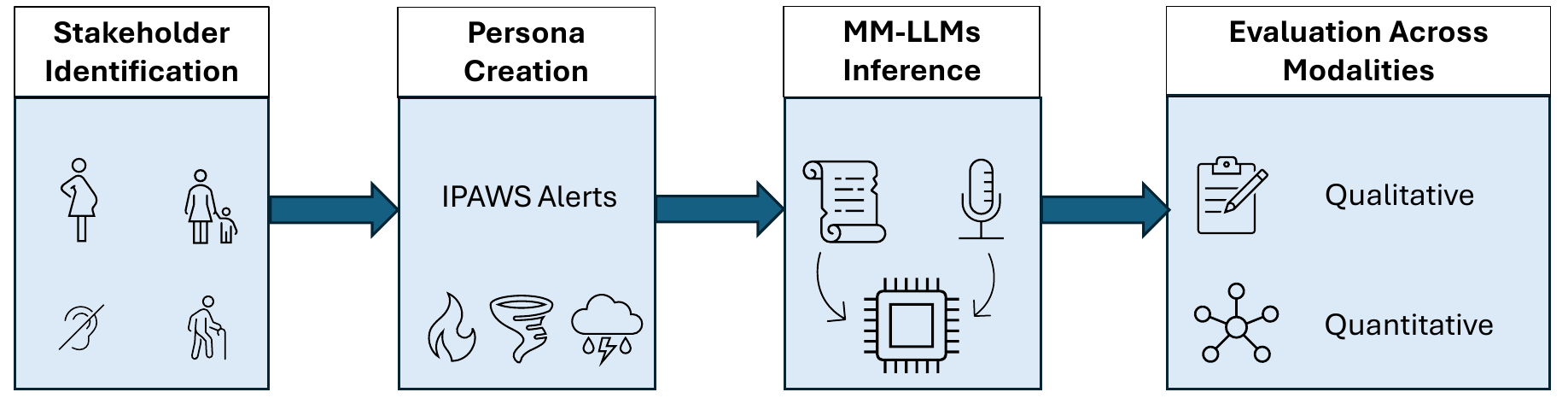}
    \caption{Framework overview illustrating the four components of: Stakeholder Identification; Persona Creation; MM-LLM Inference, and Evaluation.}
    \label{fig:framework}
\end{figure*}

Our framework (Fig.~\ref{fig:framework}) 
comprises four components: \textit{Stakeholder Identification} 
based on FEMA guidelines, \textit{Persona 
Creation} grounded in 
official 
alert types, \textit{Multimodal LLMs} that 
take these persona inputs paired as text and audio prompts, and \textit{Multifaceted Evaluation} done through manual analysis, semantic similarity metrics, and factual overlap scoring.

\subsection{Identification of Stakeholders}
To identify populations vulnerable during disaster events, we conducted a critical review of relevant resources from the U.S. Centers for Disease Control and Prevention (CDC), identifying a formal group classified as individuals with ``access and functional needs.'' According to the CDC, this category refers to individuals who may require additional assistance due to temporary conditions or permanent conditions that may limit their ability to respond effectively in emergencies~\cite{cdc2021afn}.


Notably, individuals with access and functional needs are not required to have a formal diagnosis or medical evaluation. The CDC identifies several groups that may be disproportionately affected during emergencies, including children (with or without disabilities), pregnant women, older adults, individuals with physical, sensory, intellectual, developmental, cognitive, or mental disabilities, those with chronic health conditions or pharmacological dependencies, people with limited English proficiency, and individuals facing financial, transportation, or legal barriers to emergency preparedness and recovery. From this list of individuals, we selected four stakeholder groups based on two eligibility criteria: (1) the existence of prior literature on the group and (2) high vulnerability during disaster scenarios. The final stakeholder groups considered in this study include: \textit{Pregnant Women, Mothers with Toddlers, Hard of Hearing People, Elderly Individuals with Dementia}.
Disasters have been linked to potential adverse outcomes and impacts for pregnant
women~\cite{harville2021assessing}. Their increased rates of preterm birth, pregnancy complications, and maternal mortality, with documented cases from Hurricane Katrina and other events showing disproportionate health impacts on this group~\cite{callaghan2007health}. Their mobility limitations and time-sensitive medical needs make rapid, actionable emergency communication especially critical.
It is widely recognized that emergency plans should account for the unique needs of mothers and children, with the youngest
infants being the most vulnerable to the effects of natural disasters~\cite{mudiyanselage2022infant}.
Mothers with toddlers face increased risk of their children getting acute illness, developmental disabilities, being underweight and the mothers developing maternal anxiety~\cite{yamazaki2023understanding}. The Great East Japan Earthquake and Tsunami has been described as one of the worst natural disasters in Japanese history that  affected the physical and socioenvironmental conditions of the local communities including mothers with infants and preschool-aged children~\cite{nishihara2018factors}.
Hard of hearing individuals represent one of the most directly impacted groups in 
multimodal risk communication research since the standard emergency alerting systems 
(sirens, broadcast audio, Wireless Emergency Alerts with sound) are either inaccessible or partially inaccessible to them, making alternative modalities precisely the focus of this study. This trend is illustrated by the preemptive evacuation of hard of hearing individuals during Hurricane Rita where hard of hearing individuals evacuated preemptively due to hurricane announcements being exclusively accessible through specific television stations, translators being unavailable at shelters, and information from FEMA and the Red Cross not being conveyed in sign language or any other accessible manner ~\cite{tannenbaum2024risk}, illustrating 
how communication gaps could materially alter evacuation behavior and risk exposure.
Finally, research shows 
adverse impact of disasters on memory and awareness of elderly individuals with dementia~\cite{fahmy2026effects}. Some older individuals impacted by hurricanes experienced a transient decrease in working memory lasting 6 months after the disaster, with a subsequent return to pre-disaster levels by the 14-month follow-up period~\cite{fahmy2026effects}. Similarly, those affected by the earthquake in Turkey experienced significant declines in memory, daily functioning, speech, and overall cognitive abilities~\cite{guney2026experiences}. Based on this evidence, our study scope includes four vulnerable user groups as stakeholders.  


\subsection{Persona Creation}
Based on the identified stakeholders, we create personas and communication scenarios corresponding to each. These scenarios are derived from real emergency alerts that are being sent out in coordination with the 
FEMA's national system for local alerting called the Integrated Public Alert \& Warning System (IPAWS)\footnote{\url{https://www.fema.gov/emergency-managers/practitioners/integrated-public-alert-warning-system}}, which delivers verified emergency information to the public via multiple channels, including mobile phones through Wireless Emergency Alerts (WEA)\footnote{\url{https://www.weather.gov/wrn/wea360}}, broadcast media through the Emergency Alert System, and NOAA Weather Radio. Using samples of such alerts, we constructed a list of 40 distinct prompts that reflect various alert types and disaster scenarios.

For instance, considering `\textit{Flash Flood Warning}', the prompt template is ``I am a X, and I just received a flash flood warning. What should I do during this time to stay safe?'' and similarly, when considering `\textit{Flash Flood Alert}', the prompt template is ``I am a X, and I just received a flash flood alert. What should I do during this time to stay safe?'' X represents a persona; we have summarized the four different personas, along with their sample queries, in Table~\ref{tab:prompts}. For the audio prompts one person from the team recorded their voice to generate the query which was then supplied to the MM-LLM. \textit{(Note: we will share the full set of prompts and result logs with the accepted, camera-ready paper.)}

\begin{table*}[t!]
\centering
\caption{Illustrative prompts for validating consistency across MM-LLMs including the
Wireless Emergency Alerts sent out by National Weather Service during emergency
situations.}
\label{tab:prompts}
\renewcommand{\arraystretch}{1.2}
\begin{tabular}{|p{4cm}|p{3cm}|p{10cm}|}
\hline
\textbf{Persona Name} & \textbf{Alert Type} & \textbf{Persona Query} \\
\hline
Pregnant woman & Flood alert & \textit{``I am a pregnant woman, and I just received a flood alert. What should I do during this time to stay safe?''} \\
\hline
Mother with toddler & Flood warning & \textit{``I am a mother with a toddler, and I just received a flood warning. What should I do during this time to stay safe?''} \\
\hline
Hard of hearing person & Flood warning & \textit{``I am a hard of hearing person, and I just received a flood warning. What should I do during this time to stay safe?''} \\
\hline
Elderly individual with dementia & Flood warning & \textit{``I am an elderly individual with dementia, and I just received a flood warning. What should I do during this time to stay safe?''} \\
\hline
\end{tabular}
\end{table*}

For scoping purpose, it is important to distinguish between two layers of accessibility challenge that 
MM-LLMs must address for users with access and functional needs. The first is 
\textit{perceptual accessibility} which is whether a user can physically receive and 
decode the signal delivered by an interface (e.g., whether a hard of hearing 
individual can perceive an audio alert or whether a person with dementia can 
retain spoken instructions). The second is \textit{informational consistency} which is
whether the system produces equivalent content regardless of the modality through 
which a query is submitted. This study evaluates the latter. Our methodology 
treats audio as a clean digital signal passed programmatically to the MM-LLM, 
and therefore does not simulate the perceptual barriers such as signal 
frequency, clarity, or amplification requirements that hard of hearing 
individuals face when receiving audio in real-world environments. Rather, we 
evaluate whether the model itself introduces asymmetry in the information it 
provides across modalities, which is also necessary 
to meet information-access expectations. A system that produces perfectly consistent outputs across 
text and audio modalities is still inaccessible to a hard of hearing individual 
if the audio channel itself cannot be perceived. For the current study's scope, we leave this exploration for future work, 
including realistic acoustic degradation, hearing aid signal processing simulations, and 
user studies with hard of hearing participants. 

\subsection{Multimodal LLMs: State-of-the-Art Open-Weight Models}
We evaluate three state-of-the-art open-weight MM-LLMs across 40 different prompts for the four personas summarized in Table~\ref{tab:prompts}. The MM-LLMs that we employed are:

\subsubsection{AudioFlamingo}
Audio Flamingo 3 is based on a 7B language model and the LLaVA architecture. This model is trained on a unified AF-Whisper audio encoder based on Whisper that handles understanding beyond speech recognition. Audio Flamingo 3 is able to handle three distinct signal types in audio: sound, music, and speech~\cite{goel2025audio}.

\subsubsection{Qwen}
Qwen3-Omni is a natively end-to-end, multilingual model capable of processing text, images, audio, and video while delivering real-time streaming responses in both text and natural speech. The model incorporated several architectural upgrades that significantly improved performance and efficiency across MM applications spanning audio, image, video, and audio-visual tasks~\cite{xu2025qwen3}.

\subsubsection{SALMONN}
SALMONN (Speech, Audio, Language, and Music Open Neural Network) is designed to process and reason over diverse audio inputs, including speech, environmental sounds, and music. It integrates pretrained audio encoders with a LLM 
through alignment layers, enabling capabilities such as audio captioning, speech comprehension, and audio-based reasoning. SALMONN is particularly notable for its ability to generalize across heterogeneous audio domains, making it suitable for applications that require contextual understanding of real-world audio 
signals~\cite{tang2023salmonn}.

\subsection{Multifaceted Evaluation}
In order to understand the consistency of responses across text and audio modalities, we employed existing approaches (1) Using semantic similarity metrics and (2) Factual consistency metrics to evaluate LLM consistency. For each prompt, 4 personas yielded 160 responses, given the 40 prompts in our study. This resulted in 160 responses for text and 160 responses for audio. For these response pairs, the semantic similarity scores were calculated using four existing metrics: Bert (sBert), BLEU (sBLEU), Rouge (sRouge), and USE (sUSE). We calculated sBert using the BERTScore Python package, sBLEU using the NLTK Python package, sRouge using the rouge-score Python package, and sUSE using the universal-sentence-encoder model on Kaggle~\cite{wu2025estimating}.
Inspired by FactSumm~\cite{shakil2024utilizing}, factual consistency is calculated through a factual overlap score
that measures how much content from the text 
response is preserved in the audio response. The text response is treated as the source and the audio response as the candidate, extracting named entities and disaster-relevant action terms (e.g., \textit{evacuate}, \textit{shelter}) selected based on common disaster terminology. The factual overlap score is then computed as the harmonic mean of the entity precision score, which is the proportion of candidate keywords and entities also present in the source, with the ROUGE-L lexical overlap score between the two responses. 




\section{Results and Discussion}
\label{sec:results}
As defined in Section~\ref{sec:methodology}, we evaluated semantic similarities across response pairs using previously published methods and further assessed 
factual consistency using the factual overlap metric, 
which combines 
entity-level fact overlap with lexical similarity to measure whether safety-critical 
facts present in a text response are preserved in the corresponding audio response. 
Table~\ref{tab:results} reports these metrics across the three models.

\subsection{Model Performance Analysis}
As shown in Table~\ref{tab:results}, when comparing each of the four metrics to evaluate the similarity of response pairs: Bert, BLEU, Rouge, and USE; 
Qwen achieved the strongest performance across all the semantic similarity metrics, indicating the highest level of consistency between text and audio responses. The high S-BERT score (0.896) indicates that the underlying meaning was preserved more effectively across modalities. SALMONN demonstrated moderate performance, suggesting limited overlap in wording and possible scope for improvement. Flamingo, on the other hand, demonstrated the lowest performance across all metrics, indicating weaker alignment in meaning across modalities. Among all models, S-USE and S-BERT are consistently higher than S-BLEU and S-ROUGE, indicating that while models often preserve core meaning, they 
vary substantially 
in wording and structure across modalities.

\begin{table}[h]
\caption{Performance of the three different MM-LLMs across modalities evaluated using semantic consistency metrics}
\label{tab:results}
\begin{center}
\begin{tabular}{|c|c|c|c|c|}
\hline
\textbf{Model} & \textbf{S-ROUGE} & \textbf{S-BLEU} & \textbf{S-USE} & \textbf{S-BERT} \\ 
\hline
Flamingo  & 0.2227 & 0.0652 & 0.6189 & 0.4948 \\ 
\hline
SALMONN & 0.2486 & 0.1318 & 0.6964 & 0.6253 \\ 
\hline
Qwen & 0.3408 & 0.3426 & \textbf{0.8955} & \textbf{0.8961}\\ 
\hline
\end{tabular}
\end{center}
\end{table}

\subsection{Error Analysis}
\begin{figure}[t]
    \centering
    \includegraphics[width=0.48\textwidth]{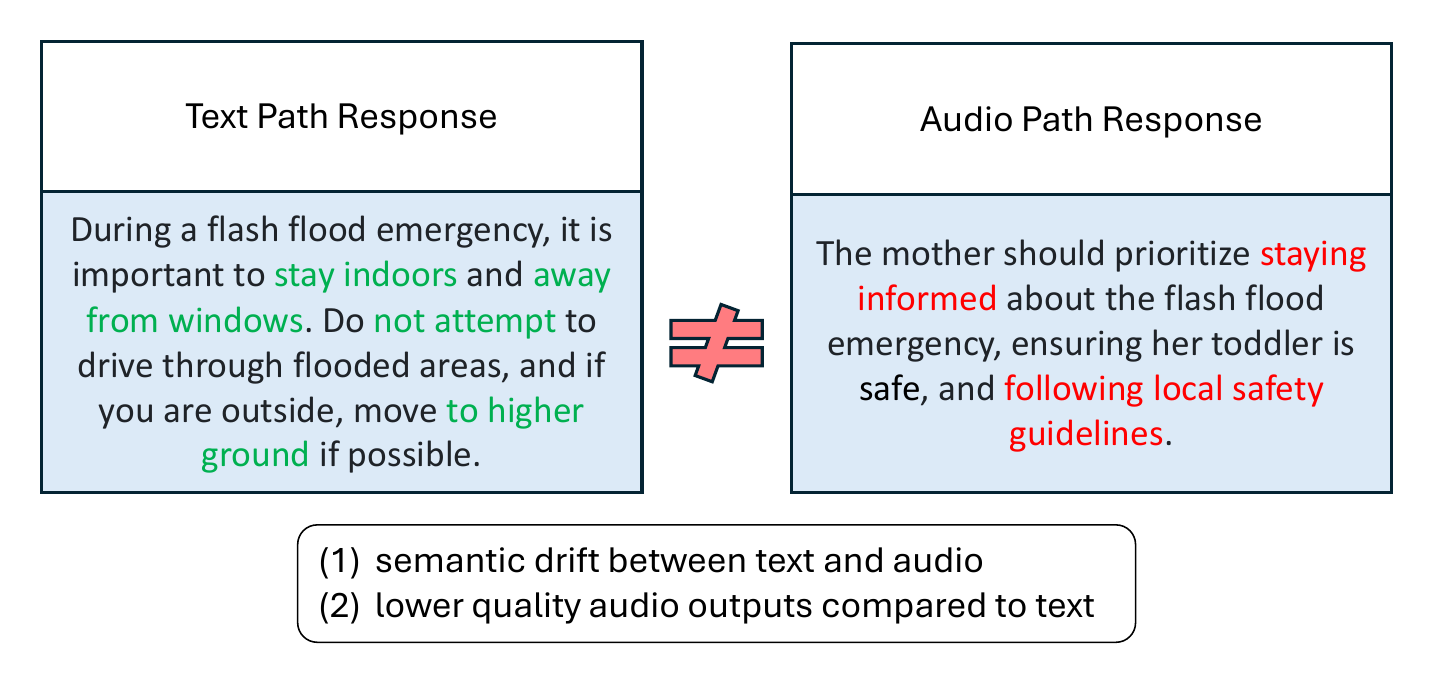}
    \caption{Cross-modal inconsistency in AudioFlamingo for a mother with a 
    toddler persona given a flash flood emergency where the text response provides 
    specific, actionable guidance (stay indoors, move to 
    higher ground), while the audio response remains vague and omits critical 
    safety instructions.}
    \label{fig:pipeline}
\end{figure}

The examples in Table~\ref{tab:error} highlight systematic inconsistencies between text and audio responses across modalities, with direct implications for the deployment of MM-LLMs in disaster-affected communities. In both cases, the text responses provide detailed, actionable guidance tailored to the specific hazard (e.g., avoiding dust exposure, staying hydrated, and limiting physical exertion), whereas the corresponding audio responses are lower quality, generic and, in  some cases, fail to address the scenario meaningfully. For instance, in the dust advisory case, the audio response omits critical protective measures such as 
\textit{filtration}, \textit{indoor sheltering}, and \textit{mask use} and instead, offers vague instructions about following safety protocols. In a humanitarian context, where affected individuals may lack access to follow-on resources, a caregiver, or reliable connectivity, such vagueness is not merely a quality issue but a safety risk. Elderly individuals with dementia, in particular, are unlikely to seek clarification or consult secondary sources under crisis situations~\cite{bell2023mortality} and the first response they receive is often the only one they act on. Another 
concern observed from these responses is the semantic misalignment across modalities as the core meaning drifts across text and audio. In the case of a mother with a toddler (see Fig.~\ref{fig:pipeline}), the audio output produces a response entirely unrelated to the user's condition, representing a major drift between the persona-adapting capabilities across both modalities. 

These patterns reflect a broader challenge in deploying AI-based communication tools for social good that the current open-weight MM-LLMs are optimized for general 
capability benchmarks rather than for the reliability that disaster risk communication contexts of underserved populations demand. 
The modality-dependent degradation observed here, where the audio interface, most likely to be used by individuals with visual impairments, low literacy, 
or motor disabilities, consistently receives less 
comprehensive guidance, 
represents a direct failure to meet that standard. Humanitarian technology 
deployments of MM-LLMs must therefore treat cross-modal consistency not as 
a secondary evaluation criterion, but as a core requirement alongside accuracy 
and factual correctness.


\begin{table}[t]
\caption{Examples of responses by 
AudioFlamingo
that demonstrated poor consistency when prompted in different modalities for the same 
query scenario. (\textcolor{blue}{Blue} highlights 
the 
situation; \textcolor{green}{Green} indicates helpful tips generated only in text modality.)}
\label{tab:error}
\begin{center}
\renewcommand{\arraystretch}{1.3}
\begin{tabular}{|p{1.8cm}|p{3.2cm}|p{2.0cm}|}
\hline
\textbf{Input} & \textbf{Text Response} & \textbf{Audio Response} \\
\hline
\textit{I am an elderly individual with dementia, and I just received a \textcolor{blue}{dust advisory}. What should I do to stay safe?} &
During a dust advisory, it's important to \textcolor{green}{stay indoors} and \textcolor{green}{avoid exposure} to dust. Keep windows and doors closed, and use a \textcolor{green}{HEPA filter} to help remove dust from the air. If you must go outside, wear a dust mask and avoid areas with high dust levels. &
The elderly individual should follow safety protocols, possibly seek guidance from a healthcare provider, and stay informed about the advisory. \\
\hline
\textit{I am a mother with toddler, and I just received an \textcolor{blue}{excessive heat warning}. What should I do to stay safe?} &
During an excessive heat warning, it's important to \textcolor{green}{stay indoors} and avoid direct sunlight. Keep hydrated by drinking \textcolor{green}{plenty of water}, and avoid strenuous activities. If possible, \textcolor{green}{use fans} or air conditioning to cool down the room. &
The speaker should follow safety protocols, possibly seek shelter, and stay informed about the situation. \\
\hline
\end{tabular}
\end{center}
\end{table}

\subsection{Modality Type Analysis}
For practical usage in risk communication where equitable access to emergency information is fundamental, 
the consistency gap revealed across 
all three models is of great concern. Based on Table~\ref{tab:results}, current open-weight MM-LLMs preserve high-level intent while failing to maintain alignment in actionable detail and semantics. Qwen's near parity between S-USE (0.896) and S-BERT (0.896) reflects relatively robust semantic and factual grounding, however, its substantially lower S-BLEU and S-ROUGE scores indicate that consistency is achieved through paraphrasing rather than faithful content reproduction. This 
matters in safety-critical scenarios where reformulation of key guidance 
such as evacuation routes or hazard avoidance steps can materially alter the usefulness of a response. On the other hand, scores from SALMONN suggest partial loss of fine-grained meaning across modalities. AudioFlamingo performs worst with both lexical divergence and semantic drift, underscoring 
the need for evaluation metrics that explicitly capture action-level agreement and factual omission. In 
risk communication, where a missed instruction to stay indoors, seek higher ground, or avoid downed power lines can directly determine survival outcomes, modality-level paraphrasing is not a stylistic concern but a humanitarian design failure that future MM-LLMs 
must explicitly address.
\subsection{Persona Type Analysis}
Table~\ref{tab:persona_results} presents factual overlap scores broken down by persona 
across all three models. The hard of hearing persona scores lowest in Qwen (0.397) and AudioFlamingo (0.237), suggesting greatest factual divergence across modalities. 
Conversely, this persona scores highest in 
SALMONN (0.306), revealing that models handle this persona differently across architectures. A contrasting pattern also appears for the 
pregnant woman persona where it is strongest persona for Qwen (0.452) and 
AudioFlamingo (0.335) but the weakest for SALMONN (0.296), indicating that 
persona-specific factual consistency is not a stable property across MM-LLMs. 
More broadly, responses involving condition-specific needs show lower factual overlap scores,
introducing a form of modality-dependent 
bias where individuals with specialized needs receive incomplete or less 
actionable information. These results show that current open-weight MM-LLMs lack robustness in handling persona-specific constraints, required for designing accessible systems 
in disaster 
contexts.
\subsection{Implications, Limitations, and Future Work}
%

These findings carry direct implications for MM-LLM based system design and humanitarian 
deployment. The 
multimodal capability of accepting diverse inputs is different from multimodal 
reliability, i.e., a system that produces inconsistent outputs across modalities provides unequal service to users whose access needs determine how they communicate, encoding modality-dependent inequity into the 
technology. Cross-modal consistency should therefore be 
treated as a priority evaluation criterion alongside accuracy, 
with training pipelines incorporating explicit consistency objectives and 
fine-tuning on  domain knowledge for relevant 
user personas  
\cite{tang2023salmonn, 
xu2025qwen3}.

At the deployment level, the persona-level variation in our results highlights 
the need for inclusive evaluation protocols that assess model behavior across the needs of
diverse users 
rather than aggregate benchmarks alone. A model that 
performs well on average but fails for hard of hearing individuals or elderly 
users with dementia is not fit for humanitarian use.

The limitations of this study point 
to at least five future directions. 
First, all audio prompts were recorded by a single human speaker, which does not capture natural variation in accent, pace, pitch, or age-related vocal characteristics that  
MM-LLMs encounter in the real world. Future work should incorporate audios generated by diverse speaker profiles and expand the prompt set to confirm whether observed patterns hold across a broader and more naturalistic range of vocal inputs. Second, the evaluation scope is relatively small and findings should be treated as indicative rather than definitive. To address this, future work should develop a framework of semantic, factual, and information-based metrics that capture safety-critical omissions in a more fine-grained manner across modalities, alongside appropriate significance tests to establish whether the reported gaps constitute robust findings.
Third, the persona-based prompting strategy assumes explicit self-disclosure of vulnerable status, which may not reflect naturalistic user interactions, and the absence of a neutral baseline condition prevents isolation of whether consistency gaps are specific to vulnerable persona conditioning or reflect general cross-modal behavior. We plan to address this through community engagement; conducting surveys and interviews with a broader set of individuals to expand, validate, and refine our persona-based scenarios with lived experience, while also establishing neutral baseline conditions in future evaluations.
Fourth, fine-tuning open-weight multimodal LLMs on disaster risk communication corpora with explicit cross-modal consistency objectives and persona-conditioned training data represents a promising path toward more equitable and reliable humanitarian AI systems.
Fifth, disaster-affected populations span diverse linguistic and cultural contexts that are often overlooked in model evaluation. Improving the performance of 
MM-LLMs in these contexts is essential to design and deploy 
accessible and equitable systems globally, 
particularly in low-resource language settings and multicultural communities.



\begin{table}[t!]
\caption{Factual overlap scores by persona across all three MM-LLMs.}
\label{tab:persona_results}
\begin{center}
\renewcommand{\arraystretch}{1.2}
\begin{tabular}{|l|c|c|c|}
\hline
\textbf{Persona} & \textbf{AudioFlamingo} & \textbf{SALMONN} & \textbf{Qwen} \\
\hline
Pregnant woman            & 0.3350 & 0.2964 & \textbf{0.4518} \\
\hline
Mother with toddler       & 0.2646 & 0.3146 & \textbf{0.4492} \\
\hline
Elderly w/ dementia       & 0.2708 & \textbf{0.3347} & 0.4258 \\
\hline
Hard of hearing           & 0.2369 & 0.3060 & \textbf{0.3970} \\
\hline
\textbf{Overall}          & 0.2765 & 0.3129 & \textbf{0.4310} \\
\hline
\end{tabular}
\end{center}
\end{table}

%


\section{Conclusion}
\label{sec:conclusion}
The primary goal of this study is to investigate the existing open-weight MM-LLMs for accessible disaster risk communication contexts. We identify stakeholders in disaster scenarios by reviewing existing literature. We then create personas and scenarios that represent risk communication situations involving the identified stakeholders. These serve as prompts in audio mode and text mode and the obtained output pairs were evaluated using semantic similarity metrics. 
The findings of this work reveal that open-weight MM-LLMs are unable to achieve complete cross-modal consistency, with performance disparities most pronounced for personas with specialized access and functional needs. These results provide evidence of existing accessibility limitations and  motivate the development of more inclusive humanitarian AI systems. 

\section{Acknowledgment}
This research was partially supported by the grant \# 2531369 
from the National Science Foundation. During the preparation of this work, the authors used Claude for Fig. 1 preparation, code assistance, and literature review support. After using this tool, the authors reviewed and edited the content as needed and take full responsibility for the content of the manuscript.

\bibliographystyle{IEEEtran}
\bibliography{bibfile}

\end{document}